\documentclass{article}
\usepackage{spconf,amsmath,graphicx}
\usepackage{blindtext}
\usepackage{multirow}
\usepackage{cleveref}

\usepackage{enumitem}
\usepackage{etoolbox}
\usepackage{graphicx}
\usepackage{makecell}
\usepackage{caption}
\usepackage{wrapfig}
\usepackage{setspace}
\usepackage{subcaption}
\usepackage{floatflt}
\usepackage{float}
\usepackage{setspace}

\DeclareMathOperator*{\argmax}{arg\,max}

\title{Investigating the Effect of Language Models in Sequence Discriminative Training for Neural Transducers}
\name{Zijian Yang, Wei Zhou, Ralf Schlüter, Hermann Ney
\vspace{-2mm}
}
\address{Machine Learning and Human Language Technology, Computer Science Department,\\
RWTH Aachen University, 52074 Aachen, Germany,\\ AppTek GmbH, 52062 Aachen, Germany
\vspace{-3mm}
}
\begin{document}
\copyrightnotice{979-8-3503-0689-7/23/\$31.00~\copyright2023 IEEE}
%
\maketitle
\begin{abstract}
In this work, we investigate the effect of language models (LMs) with different context lengths and label units (phoneme vs. word) used in sequence discriminative training for phoneme-based neural transducers. Both lattice-free and N-best-list approaches are examined. For lattice-free methods with phoneme-level LMs, we propose a method to approximate the context history to employ LMs with full-context dependency. This approximation can be extended to arbitrary context length and enables the usage of word-level LMs in lattice-free methods. Moreover, a systematic comparison is conducted across lattice-free and N-best-list-based methods. Experimental results on Librispeech show that using the word-level LM in training outperforms the phoneme-level LM. Besides, we find that the context size of the LM used for probability computation has a limited effect on performance. Moreover, our results reveal the pivotal importance of the hypothesis space quality in sequence discriminative training.
\end{abstract}
\begin{keywords}
Speech recognition, sequence discriminative training, neural transducer
\end{keywords}
\section{Introduction \& Related works}
\label{sec:intro}

In recent years, there has been significant progress in the field of automatic speech recognition (ASR) with the sequence-to-sequence (seq2seq) modeling methods. These approaches have demonstrated remarkable success in capturing the sequential nature of speech data. 
Notably, the attention-based encoder-decoder (AED) model \cite{bahdanau2016end, chan2016listen, Tske2020SingleHA}, connectionist temporal classification (CTC) \cite{graves2006connectionist} and recurrent neural network transducer (RNN-T) \cite{graves2012sequence} have become the major trends. Among these methods, RNN-T has obtained significant interest due to its ability to handle streaming tasks while maintaining competitive performance \cite{rao2017exploring}.

Sequence discriminative training criteria have demonstrated their effectiveness in improving the performance of ASR models \cite{vesely2013sequence, shannon17b_interspeech}. Prominent criteria encompass maximum mutual information (MMI) \cite{bahl1986maximum}, boosted MMI (bMMI) \cite{povey2008boosted}, and minimum Bayes risk training (MBR) criteria including minimum phone error (MPE) \cite{vesely2013sequence,povey2005discriminative}, minimum word error rate (MWE) \cite{vesely2013sequence,povey2005discriminative,Guo2020EfficientMW, prabhavalkar2018minimum} and state-level minimum Bayes risk (sMBR) training \cite{vesely2013sequence, senior2015acoustic}. Typically, these methods necessitate decoding during training to generate lattices or N-best lists as the discriminative hypothesis space, which is time-consuming and resource-intensive. 
To improve the training efficiency, \cite{povey2016purely} introduced lattice-free MMI (LF-MMI) for hybrid models, eliminating the need for the decoding step. 
Subsequently, other lattice-free (LF) methods like LF-sMBR \cite{kanda2018lattice, Michel2019ComparisonOL} and LF-bMMI \cite{zhang2021lattice} were proposed for hybrid/CTC models. LF-MMI was also utilized for AED and RNN-T models \cite{tian2022consistent, tian2022integrating} as an auxiliary loss on the encoder output. Instead of utilizing LF criteria as an auxiliary loss on the encoder output, \cite{yang2023lattice} proposed various LF objectives for phoneme-based neural transducers, applied to the final posterior output of neural transducers with limited context dependency. 
Despite achieving a significant speedup, a performance gap is observed between LF methods with a bigram phoneme-level LM and the N-best-list-based MBR  criterion with a 4-gram word-level LM.
By delving into this aspect, we seek to bridge the performance gap and advance the understanding of employing LMs in the context of sequence discriminative training. The motivation for our work begins from here. 

In this paper, we examine the effect of LMs with different context lengths and label units (phoneme vs. word) used in sequence discriminative training for phoneme-based neural transducers.
We conduct systematic comparisons crossing N-best-list based MMI and MBR training, as well as LF-MMI training.
To employ a higher-order phoneme-level LM in an LF manner, we propose a method to approximate the label history during recombination. 
The proposed method can be extended to arbitrary context length and enables the usage of word-level LM with limited context length in LF training.
Experiments under various settings on the Librispeech corpus \cite{panayotov2015librispeech} consistently show that 
\begin{itemize}[itemsep=-0.3mm]
    \item Word-level LM performs better than phoneme-level LM when a good hypothesis space is applied.
    \item The context size of the LM used for probability computation in training is not critical
    \item The quality of the hypothesis space for sequence discriminative training is critical for performance. When the hypothesis space is of good quality, its size becomes less important, i.e. a small N-best list is sufficient. When the hypothesis space is suboptimal, the biased error patterns that occur in the hypothesis space can lead to biased model outputs during sequence discriminative training, resulting in suboptimal performance.
\end{itemize}

\section{Phoneme-Based Transducers}
\label{sec:phoneme_transducer}

Following \cite{yang2023lattice, zhou2021phoneme}, the strictly monotonic topology is utilized for neural transducers. Let $X$ be the input sequence and $h_1^T$ be the hidden states generated by the encoder. The posterior probability of the output phoneme label sequence $a_1^S$ is computed by the following equation.

\begin{align*}
    \begin{split}
        P_\text{RNNT}(a_1^S|X) &= \sum_{y_1^T: \mathcal{B}(y_1^T)=a_1^S} P_\text{RNNT}(y_1^T|h_1^T)\\
        &= \sum_{y_1^T: \mathcal{B}(y_1^T)=a_1^S} \prod_{t=1}^T P_\text{RNNT}(y_t|\mathcal{B}(y_1^{t-1}), h_t)
    \end{split}
\end{align*}
Here $y_1^T$ is the alignment sequence with phoneme labels and blanks $\epsilon$, $\mathcal{B}$ is a collapse function that maps the alignment to the label sequence by removing blanks. According to \cite{zhou2021phoneme, Prabhavalkar2021LessIM, Ghodis20}, 
the model can be simplified with a limited context dependency based on a $k_\text{th}$ order Markov assumption.
\begin{equation*}
    P_\text{RNNT}(y_t|\mathcal{B}(y_1^{t-1}), h_t) = P_\text{RNNT}(y_t|a_{s_{t-1}-k+1}^{s_{t-1}}, h_t)
\end{equation*}
Here $s_t$ indicates the position in $a_1^S$ that $y_t$ reaches. $k$ is the context size. Given a training pair $(X, a_1^S)$, the sequence-level cross-entropy (CE) objective is employed for training:
\begin{equation*}
\notag
    \mathcal{L}_\text{CE} = - \log P_\text{RNNT}(a_1^S|X)
\end{equation*}
During decoding, the decision rule is formulated as follows:
\begin{equation*}
  X \rightarrow \mathcal{W}({a^*}_1^{S^*}) = \argmax_{S, \mathcal{W}(a_1^S)} \big[ P_\text{RNNT}(a_1^S|X) \cdot \frac{P^{\lambda_1}_\text{ELM}(\mathcal{W}(a_1^S))}{P^{\lambda_2}_\text{ILM}(a_1^S)} \big]
\end{equation*}
Here $\mathcal{W}$ is the lexical mapping function converting the output phoneme sequence into the corresponding word sequence. We also incorporate an external LM and internal LM (ILM), denoted as $P_\text{ELM}$ and $P_\text{ILM}$, respectively. Two scales, $\lambda_1$ and $\lambda_2$, are introduced for LM integration.

\section{MMI Training}


The MMI objective is defined as follows: 
\begin{equation}
    \mathcal{L}_\text{MMI} = -\log \frac{q_\text{seq}(a_1^S|X)}{\sum_{S',{a'}_1^{S'}}q_\text{seq}({a'}_1^{S'}|X)}
    \label{eq:MMI}
\end{equation}
where $q_\text{seq}(a_1^S|X)$ is the combined probability of RNN-T and LM probabilities for a given phoneme sequence $a_1^S$.

\begin{equation*}
    q_\text{seq}(a_1^S|X) = P_\text{RNNT}(a_1^S|X) \cdot P_\text{LM}(a_1^S)
\end{equation*}

The numerator in \Cref{eq:MMI} is equivalent to a standard CE loss. The denominator, which contains a summation over all possible sequences, is usually approximated by an N-best list/lattice or computed directly in an LF manner.
For the latter, the computation can be done using dynamic programming (DP) with different approximations/model assumptions for the recombination process.
We utilize the phoneme transducer with context one as our acoustic model (AM). In the following discussion, we also use phonemes as the basic unit.

\subsection{LF-MMI Training with Phoneme-level LMs}
When using a phoneme-level LM, as the label units of the LM and AM are the same, it is efficient to apply the LF method to compute the denominator. In this section, LF methods for LMs with different context lengths are introduced. To enable the usage of LMs with full-context dependency, we propose a method to approximate the context of the LM. 

\subsubsection{Phoneme-level LMs with Limited Context Dependency}
 When the context length of the LM is limited to $k$, the recombination becomes feasible for the same limited history $u_1^{k} \in \mathcal{V}^k$. As discussed in \cite{yang2023lattice}, the summation in the denominator can be efficiently computed by DP.\\
\scalebox{0.9}{\parbox{1.1\linewidth}{
\begin{equation*}
    \sum_{S',{a'}_1^{S'}}q_\text{seq}({a'}_1^{S'}|X) = \sum_{u_1^k} Q_\text{MMI}(T, u_1^k)
\end{equation*}}}
Here the auxiliary function $Q_\text{MMI}(t,u_1^k)$ is defined as:\\
\scalebox{0.9}{\parbox{1.1\linewidth}{
\begin{equation}
Q_\text{MMI}(t,u_1^k) = \sum_{s,a_1^s:a_{s-k+1}^s=u_1^k} q_\text{seq}(a_1^s|X,t)
\label{eq:mmiauxi}
\end{equation}}}
where $q_\text{seq}(a_1^s|X,t)$ is the partial probability of the sub-sequence $a_1^s$ up to time frame $t$. 
\begin{align*}
    q_\text{seq}(a_1^s|X,t) &=  \sum_{y_1^t: \mathcal{B}(y_1^t)=a_1^s}q_\text{seq}(y_1^t|X)\\
    &= \sum_{y_1^t: \mathcal{B}(y_1^t)=a_1^s} \prod_{\tau=1}^t q_\text{seq}(y_{\tau}|a_{s_{\tau-1}-k+1}^{s_{\tau-1}},\tau)
\end{align*}
Here $q_\text{seq}(y_{\tau}|a_{s_{\tau-1}-k+1}^{s_{\tau-1}}, \tau)$ is denoted as the combined probability of the RNN-T and LM output at position $(\tau, y_{\tau})$. Different scales $\alpha$ and $\beta$ can be introduced for the RNN-T and LM outputs.\\
\scalebox{0.9}{\parbox{1.1\linewidth}{
\begin{align}
    \begin{cases}
    P^\alpha_\text{RNNT}(\epsilon|a_{s_{\tau-1}-k+1}^{s_{\tau-1}},h_{\tau}), &y_{\tau}=\epsilon \\
    P^\alpha_\text{RNNT}(a|a_{s_{\tau-1}-k+1}^{s_{\tau-1}},h_{\tau}) \cdot P^\beta_\text{LM}(a|a_{s_{\tau-1}-k+1}^{s_{\tau-1}}), &y_{\tau} = a \in \mathcal{V}
    \end{cases}
    \label{eq:scorecombine}
\end{align}}}
Then \Cref{eq:mmiauxi} can be computed by the DP recursion.
\scalebox{0.9}{\parbox{1.1\linewidth}{
\begin{equation}
\begin{aligned}
    Q_\text{MMI}(t,u_1^k) &= Q_\text{MMI}(t-1,u_1^k) \cdot q_\text{seq}(\epsilon|u_1^k,t)\\
            +&  \sum_{u_0} Q_\text{MMI}(t-1, u_0^{k-1}) \cdot q_\text{seq}(u_k|u_0^{k-1}, t)
\end{aligned}
\label{eq:DP_MMI}
\end{equation}}}
\vspace{-5mm}

\subsubsection{Phoneme-Level LMs with Full-Context Dependency}
\label{sec:contextapprox}
When the model has a full-context dependency, it is infeasible to recombine all sequences when the exact context is considered. Inspired by \cite{wynands2022efficient}, we apply an approximation for the context at each recombination step.
By employing this approximated context, we can efficiently perform recombination in a limited context dependency manner as introduced in \Cref{eq:DP_MMI}. 
Given the context limit $k$ for recombination, for each pair $(t,u_1^k)$, a full context $\Tilde{C}(t,u_1^k)$ is applied as an approximated context. In the following discussion, we will introduce how $\Tilde{C}(t,u_1^k)$ is obtained iteratively during the time-synchronized recombination process. To simplify the notations, given a fixed context $u_1^k$, we denote $\omega(t, \epsilon) = Q_\text{MMI}(t-1,u_1^k) \cdot q_\text{seq}(\epsilon|u_1^k, t)$ and $\omega(t, u_0) = Q_\text{MMI}(t-1, u_0^{k-1}) \cdot q_\text{seq}(u_k|u_0^{k-1}, t)$ as scores for blank and phoneme labels correspondingly.
The method of generating the approximated context is described below. Firstly, we find out the best predecessor hypothesis $\Tilde{u}_t$ over phoneme labels in $\mathcal{V}$ and $\epsilon$.
\begin{align*}
\begin{split}
    \Tilde{u}_t &= \argmax_{ u \in \mathcal{V} \cup \{\epsilon\}} \omega(t, u)
\end{split}
\end{align*}
Then, we update the context by the following equation.
\begin{align*}
    \Tilde{C}(t,u_1^k) = \Bar{C}(t, \Tilde{u}_t)
\end{align*}
where $\Bar{C}(t, \Tilde{u}_t)$ is the updated context given label or blank as the next input.
\begin{align*}
    \Bar{C}(t, \Tilde{u}_t) = \begin{cases}
        \Tilde{C}(t-1, u_1^k),& \Tilde{u}_t = \epsilon \\
        \Tilde{C}(t-1, u_0^{k-1}: u_0=\Tilde{u}_t) + u_k,& \Tilde{u}_t \in \mathcal{V}
    \end{cases}
\end{align*}
Here the $+$ operation on the label sequence means $u_k$ is concatenated to the previous context. If $\Tilde{u}_t = \epsilon$, the blank transition is dominant in the recombination. In this case, the context remains unchanged. If $\Tilde{u}_t \in \mathcal{V}$, the context $\Tilde{C}(t-1, u_0^{k-1})$ with the highest score is selected and updated by concatenating it with the new label $u_k$.
It is crucial to consider the score of the blank label and have the possibility to keep the context unchanged. Otherwise, the actual length of the context would be the same as the current time $t$, which is not optimal.

In this work, we employ an LSTM network to model the full-context dependency during training. Given the full context $a_1^{s-1}$, the LM probability is computed by:
\begin{equation*}
    P_\text{LM}(a_s | a_1^{s-1}) = P_\text{LSTM}(a_s | \mathbf{S}(a_1^{s-1}))
\end{equation*}
Here $\mathbf{S}$ is the hidden state of the LSTM network taking $a_1^{s-1}$ as input. With the context approximation, the combined score for a phoneme label $u_k$ at time frame $t$ is computed by:\\
\scalebox{0.88}{\parbox{1.1\linewidth}{
\begin{align*}
    q_\text{seq}(u_k|u_0^{k-1},t) = P^\alpha_\text{RNNT}(u_k|u_0^{k-1},h_t) \cdot P^\beta_\text{LM}(u_k|  \Tilde{C}(t-1,u_0^{k-1}))
    \label{eq:maxappro}
\end{align*}}}
The probability for blank is computed in the same way as in \Cref{eq:scorecombine} because the LM score is not involved. It is worth noting that the LM probability is also a function of time frame $t$ implicitly due to the varying nature of the approximated context at different time frames.


\subsubsection{Reduced Hypothesis Space}
In LF methods, typically, all possible sequences are considered in the denominator, resulting in a very large hypothesis space. This extensive space is necessary for from-scratch training, because the model cannot determine which hypotheses are the good ones and should be included in the denominator. However, for a well-initialized model, it remains uncertain how much benefit can be gained by utilizing the entire sequence space. To address this question, we reduce the size of the hypothesis space by selecting the top $J$ candidates at each time frame during the DP computation for the denominator. At each time step, only a set of top $J$ candidates $ \mathcal{U}_J(t)$ is reserved and all other candidates are pruned out.
\begin{equation*}
    \mathcal{U}_J(t) = \{u_1^k: Q(t, u_1^k) \in \text{top $J$ best}\}
\end{equation*}
When the top $J$ approximation is applied, the summation is over the candidates in $\mathcal{U}_J(t)$ rather than all $k$-grams. It is worth noting that even if only $J$ hypotheses remain at each time frame $t$, because of the recombination step, there are in total $\mathcal{O}(J^T)$ alignments considered in the denominator.

\subsection{LF-MMI Training with Word-level LMs}
Compared to the phoneme-level LM, incorporating a word-level LM with the phoneme-based transducer is less straightforward in the LF training.
In this section, we introduce two methods to compute the denominator with word-level LMs integrated.
To reduce the computation complexity, we consider all possible phoneme sequences for the generation of the hypothesis space rather than constructing a lexicon prefix tree. 
This is simply achieved by utilizing the end-of-word (EOW) augmented phoneme set \cite{zhou2021phoneme}.
More precisely, we do a pronunciation-to-word mapping whenever an EOW phoneme is emitted and apply the corresponding word-level LM score, where the best LM score is used in the homophone case. Since there is no constraint on the context, a phoneme sequence may not be assigned to any word in the lexicon. In this case, we use the probability of the unknown token.

\subsubsection{Context Approximation}
Even if the word-level context length is limited, the context dependency can still be too long when mapped to the phoneme sequence. To make the LF methods feasible, the context approximation introduced in Section \ref{sec:contextapprox} can be employed to further limit the context length for the recombination step. When the output phoneme $u_k$ belongs to the EOW phoneme set $\mathcal{E}$, the approximated context $\Tilde{C}(t-1,u_0^{k-1})$ is applied for word-level LM probability computation. However, for a word-level LM, the output distribution and context dependency are both on the word level. Therefore, the context $\Tilde{C}(t-1,u_0^{k-1})$ is separated into two parts: $\Tilde{C}_h(t-1,u_0^{k-1})$ is the phoneme history up to the last EOW phoneme contained in $\Tilde{C}(t-1,u_0^{k-1})$, and $\Tilde{C}_w(t-1,u_0^{k-1})$ denotes the current within-word context after the last EOW phoneme in $\Tilde{C}(t-1,u_0^{k-1})$. Then, the two parts are mapped to words with a function $\mathcal{W}$, employed as the word-level context and word-level output label, respectively.
The LM probability $P_\text{LM}(u_k|\Tilde{C}(t-1,u_0^{k-1}))$ is defined as follows:\\
\scalebox{0.9}{\parbox{1.1\linewidth}{
\begin{align*}
    \begin{cases}
    P_{LM}(\mathcal{W}(\Tilde{C}_w(t-1,u_0^{k-1})+u_k))| \mathcal{W}(\Tilde{C}_h(t-1,u_0^{k-1})), & u \in \mathcal{E}\\
    1, & \text{others}
    \end{cases}
\end{align*}}}
To apply the LM on word-level, the LM probability is only considered when emitting EOW phonemes, while the probability is set to be $1$ in other cases.
\subsubsection{Pruning-Recombination Approach}
\label{sec:prunerecombine}
In addition to using the approximated context, we introduce a pruning-recombination method for on-the-fly denominator computation via beam search. This approach is similar to the conventional beam search conducted on the alignment sequences $y_1^T$. However, instead of considering the individual scores of the top candidates, we perform a summation of the probabilities of all candidates in the beam that share the same context dependency as the recombination step, which increases the probability mass in the denominator. The summation is done for both within-word and word-boundary recombination at each time frame. Inspired by \cite{hori2017multi, wang2020investigation}, we also employ multi-level LM decoding in the search process. That is, before reaching the EOW phoneme, a bigram phoneme-level LM is used for search. When emitting the EOW phoneme, the probability of the word-level LM is applied, and the accumulated within-word phoneme-level LM probability is disregarded. We refer the readers to \cite{hori2017multi, wang2020investigation} for more details.

\subsection{N-Best-List Approach}
Another popular method to compute the denominator involves approximating the hypothesis space with N-best hypotheses. In this approach, N hypotheses with the best scores are selected to be the hypothesis space. To make the training stable, the reference sequence is always included in the N-best list. As $N$ is usually small, the recombination step becomes unnecessary. This allows for the direct use of LMs with varying context lengths and label units during training, as well as different cost functions used for MBR training. Besides, the LM used for N-best list generation and training can be different, introducing more flexibility. Even though the number of hypotheses used in denominator computation is limited, employing a strong LM for search ensures their high quality. However, if the LM is less powerful, it is more prone to generating bad hypotheses. When the hypothesis space is small, it may introduce biased error patterns in hypotheses.
This can result in reduced improvement or even degradation of the overall system performance. 


\vspace{-3mm}
\section{Experimental Setup}
In our experiments, we evaluate our approach on the 960-hour Librispeech dataset (LBS) \cite{panayotov2015librispeech}. The architecture of our transducer model follows \cite{yang2023lattice, zhou2022efficient}. The encoder comprises 12 conformer layers \cite{Gulati2020ConformerCT}, while the prediction network consists of 2 feed-forward layers.

For the phoneme-level LMs used in training, the 1-gram phoneme-level LM is a count-based model. As there are only 79 label units and all of them are observed during training, no smoothing method is applied. The 2-gram and 3-gram phoneme-level LMs are modeled by feed-forward neural networks with the same architecture as the prediction network, followed by a softmax layer.
The full-context phoneme-level LM is modeled by one LSTM layer, followed by a projection layer and a softmax layer. For the word-level LMs used in training, we apply 1-gram and 4-gram counte-based LMs. Word-level LMs are trained on all text data to capture the underlying word distribution accurately.

During training, we follow the 3-stage pipeline proposed in \cite{zhou2022efficient}. After the standard CE training for the whole network, we apply our sequence discriminative training methods to the well-initialized model. All the hyperparameters are tuned on the dev set. $\alpha$ and $\beta$ are individually optimized for different methods. The size of the N-best list is 4. The N-best list is generated offline with the baseline RNN-T model and 1-gram or 4-gram word-level LM. During training, the N-best list remains static. In recognition, if not specified, a word-level transformer LM following the setup in \cite{Irie2019LanguageMW} is used as ELM. We estimate the ILM by the zero-encoder approach \cite{variani2020hybrid, meng2021internal, zhou2022language}.  We use RASR2 \cite{zhou2023rasr2} to run recognition and jointly with RETURNN \cite{zeyer2018returnn} to run training.

\section{Results}







\subsection{LF-MMI Training with Phoneme-Level LMs}

\begin{table}[t!]
\caption{WER [\%] results of LF-MMI training using full-context (LSTM) phoneme LM with different top $J$ approximation. Recognition on LBS dev-other uses word-level transformer LM as ELM.} 
\centering
\begin{tabular}{|c|c|c|}
\hline
   $J$ & dev-other \\ \hline
1  &    diverge                 \\ \hline
5 &      3.8                \\ \hline
10 &     3.8                \\ \hline
20 &      3.7                \\ \hline
50 &      3.7                \\ \hline
all &     3.7                 \\ \hline
\end{tabular}
\vspace{-3mm}
\label{tab:approx_topk}
\end{table}
For the LF-MMI training with a context approximation proposed in \ref{sec:contextapprox}, we first tune the size of the hypothesis space. Table \ref{tab:approx_topk} shows the effect of the size of the hypothesis space with different top $J$ approximation. In these experiments, we use $\alpha=1.0$ and $\beta=0.2$. The phoneme-level LSTM LM is used as the full-context LM in training. As shown in Table \ref{tab:approx_topk}, when $J=1$, the denominator considers only one alignment path, providing insufficient discriminative information and leading to divergence. As $J$ grows larger, the model converges and obtains slightly better performance. However, when $J$ is larger than 20, no further improvement is observed, which shows that using the whole hypothesis space does not necessarily result in better performance for a well-initialized model. Therefore, we use $J=20$ for context approximation methods in the following experiments.

Table \ref{tab:phonlf} exhibits the results of LF-MMI training using phoneme-level LMs with different context lengths. Perplexity (PPL) of the phoneme-level LM is computed on a small cross-validation dataset with about 205k phonemes. For LMs with limited context, $\alpha=1.2$ and $\beta=0.3$, and for the LM with full context, $\alpha=1.0$ and $\beta=0.2$.
Compared to the CE baseline, all of the LF-MMI methods show improvements on dev-other and test-other datasets. With the increasing context size of the LM, PPL goes down, indicating better modeling of the phoneme prior distribution. However, in terms of WER, slight improvements are observed up to 2-gram LM and more context beyond that does not yield additional improvements, which indicates a limited effect of the LM context length.

\begin{table}[t!]
\centering
\caption{Comparison of LF-MMI training using phoneme-level LMs with different context lengths. PPLs of phoneme LMs are computed on a small cross-validation set. WERs on LBS use word-level transformer LM as ELM for recognition.}
\setlength{\tabcolsep}{0.2em}
\begin{tabular}{|c|c|c|cccc|}
\hline
\multirow{3}{*}{Criterion} & \multirow{3}{*}{phoneme LM} & \multirow{3}{*}{PPL} & \multicolumn{4}{c|}{WER {[}\%{]}}                                                                     \\ \cline{4-7} 
                           &                             &                      & \multicolumn{2}{c|}{dev}                                & \multicolumn{2}{c|}{test}          \\ \cline{4-7} 
                           &                             &                      & \multicolumn{1}{c|}{clean} & \multicolumn{1}{c|}{other} & \multicolumn{1}{c|}{clean} & other \\ \hline
CE                         & -                           &     -                 & \multicolumn{1}{c|}{1.8}      & \multicolumn{1}{c|}{3.9}      & \multicolumn{1}{c|}{2.1}      &   4.6    \\ \hline
\multirow{5}{*}{LF-MMI}    & 0-gram                          &     79                 & \multicolumn{1}{c|}{1.8}      & \multicolumn{1}{c|}{3.8}      & \multicolumn{1}{c|}{2.1}      &   4.4    \\ \cline{2-7} 
                           & 1-gram                           &     45                 & \multicolumn{1}{c|}{1.8}      & \multicolumn{1}{c|}{3.8}      & \multicolumn{1}{c|}{2.1}      &    4.3   \\ \cline{2-7} 
                           & 2-gram                           &      16                & \multicolumn{1}{c|}{1.7}      & \multicolumn{1}{c|}{3.7}      & \multicolumn{1}{c|}{2.1}      &    4.3   \\ \cline{2-7} 
                           & 3-gram                           &       10               & \multicolumn{1}{c|}{1.7}      & \multicolumn{1}{c|}{3.7}      & \multicolumn{1}{c|}{2.1}      &    4.3   \\ \cline{2-7} 
                           & full-context                        &        5              & \multicolumn{1}{c|}{1.7}      & \multicolumn{1}{c|}{3.7}      & \multicolumn{1}{c|}{2.1}      &    4.3   \\ \hline
\end{tabular}

\label{tab:phonlf}
\end{table}
\subsection{N-best-list Approach for Phoneme/Word-Level LMs}

\begin{table}[t!]
\centering
\caption{WERs [\%] for N-best-list-based approaches using different LMs for training. The N-best list is generated with a 4-gram word-level LM and remains static. Recognition on LBS uses word-level transformer LM as ELM.}
\label{tab:nbestphonword} 
\setlength{\tabcolsep}{0.2em}
\begin{tabular}{|c|c|c|cccc|}
\hline
\multirow{2}{*}{Criterion} & \multicolumn{2}{c|}{Training LM} & \multicolumn{2}{c|}{dev}                                & \multicolumn{2}{c|}{test}          \\ \cline{2-7} 
                           & unit & order & \multicolumn{1}{c|}{clean} & \multicolumn{1}{c|}{other} & \multicolumn{1}{c|}{clean} & other \\ \hline
\multirow{4}{*}{MMI}       & \multirow{2}{*}{phoneme}  & 2                               & \multicolumn{1}{c|}{1.7}      & \multicolumn{1}{c|}{3.7}      & \multicolumn{1}{c|}{2.1}      &   4.3    \\ \cline{3-7} 
                           &                           & full                   & \multicolumn{1}{c|}{1.7}      & \multicolumn{1}{c|}{3.7}      & \multicolumn{1}{c|}{2.1}      &   4.3    \\ \cline{2-7} 
                           &    \multirow{2}{*}{word}                       & 1                   & \multicolumn{1}{c|}{1.7}      & \multicolumn{1}{c|}{3.6}      & \multicolumn{1}{c|}{2.1}      &   4.2    \\ \cline{3-7} 
                           &                       & 4                               & \multicolumn{1}{c|}{1.7}      & \multicolumn{1}{c|}{3.5}      & \multicolumn{1}{c|}{2.1}      &      4.1 \\ \hline
                           
\multirow{2}{*}{MBR}       & {phoneme}  & 2                               & \multicolumn{1}{c|}{1.7}      & \multicolumn{1}{c|}{3.7}      & \multicolumn{1}{c|}{2.1}      &     4.3  \\ \cline{2-7} 
                           & word                      & 4                               & \multicolumn{1}{c|}{1.7}      & \multicolumn{1}{c|}{3.7}      & \multicolumn{1}{c|}{2.1}      &    4.1   \\ \hline
\end{tabular}
\end{table}
Table \ref{tab:nbestphonword} demonstrates the performance of the N-best-list approaches with various LMs used for probability computation in training. The N-best list is generated with a 4-gram word-level LM, and all experiments in the table share the same N-best list. When using the same phoneme-level LMs for training, compared to results in Table \ref{tab:phonlf}, it is shown that even with a relatively small hypothesis space, a good selection can yield performance comparable to using a much larger hypothesis space. 
Both phoneme-level and word-level LMs show only marginal improvement from longer contexts, which further validates the limited effect of the LM context length. When comparing the MMI and MBR criteria, we observe comparable performance across different LMs. Regardless of the context length, using word-level LMs is consistently better than phoneme-level LMs. Even using a 1-gram word-level LM, whose context length is effectively shorter than the phoneme-level LSTM LM, the performance is still slightly better, indicating that word-level statistics bring more improvement.

\subsection{Hypothesis space generation using word-level LMs}
\label{sec:wordLMs}

Table \ref{tab:wordhypgen} compares the results of MMI training using different approaches to generate the hypothesis space. The word-level LMs used for hypothesis space generation and for probability computation in training can be different.
Prune-recomb denotes the method proposed in \ref{sec:prunerecombine}, while prune-single denotes a similar process but without within-word recombination. The beam size for both methods is $20$. For the LF method with context approximation, the top $J$ approximation is used to reduce memory usage, where $J=20$. As shown in the table, when the hypothesis space is good, i.e. generated by the 4-gram LM, using a 4-gram LM provides marginal improvement compared to using a 1-gram LM. However, comparing the models using 1-gram and 4-gram in hypothesis generation, a large gap can be observed, highlighting the crucial role of hypothesis space quality in performance.

To reflect the quality of hypothesis space, we also show the recognition results using the CE baseline model with 1-gram and 4-gram word-level LMs, which is the same setup for the above hypothesis space generation.
The results on LBS dev-other are shown in Table \ref{tab:unidecode}.
When using 1-gram LM, the total WER is much worse than using the 4-gram LM, indicating that the generated hypothesis space is much worse. Moreover, we observe a high deletion error when decoding with 1-gram, denoting the generated hypotheses are generally short. This can lead to a bias in the error pattern in the hypothesis space. During sequence discriminative training, the probabilities of shorter sequences are suppressed, and the probabilities of longer sequences are over-boosted, which leads to higher substitution and insertion errors, as shown in Table \ref{tab:wordhypgen}. 
Comparing different approaches used for hypothesis generation with the 1-gram LM, pruning-recombination obtains better performance due to the phoneme-level LM used to help the search process, as well as a larger probability mass covered. When emitting an EOW phoneme, the removed within-word phoneme-level LM score compensates for the word-level LM score, which makes the score for the EOW phoneme higher. A higher number of EOW phonemes in the beam facilitates the initiation of new words during the decoding, which mitigates the high deletion error problem in the hypothesis generation process, leading to a lower insertion error after MMI training. As can be seen in the table, pruning-recombination has the lowest number of deletion errors compared to other methods when using the 1-gram LM for hypothesis space generation. For the LF context recombination method, although a larger hypothesis space is used for the denominator computation, there is no improvement obtained compared to the N-best-list method. The potential reason can be that the poor performance of the 1-gram word-level LM leads to a suboptimal context approximation.

Table \ref{tab:wordlfnbest} shows similar conclusions with a stronger LM used for recognition, i.e. the transformer LM. Employing the pruning-recombination method marginally outperforms the N-best-list approach on the test-other dataset when the 1-gram LM is used for both hypothesis generation and probability computation. When using 4-gram LM for hypothesis space generation, the gap between 1-gram and 4-gram LM used for training is marginal, while the difference between using 1-gram and 4-gram LMs for hypothesis space generation is again rather large.
This further verifies the importance of the hypothesis space quality.  
\begin{table}[]
\caption{WER [\%] results of MMI training using different word LMs and hypothesis space generation methods. Note that the word LM for probability computation in training can be different than the one used for hyp-space generation. Recognition on LBS dev-other uses 4-gram word-level LM.}
\setlength{\tabcolsep}{0.28em}
\begin{tabular}{|c|c|c|c|c|c|c|}
\hline
\multicolumn{2}{|c|}{Hyp-Space Generation} & Training & \multicolumn{4}{c|}{dev-other}                                                         \\ \cline{1-2} \cline{4-7} 
Method & LM & LM & \multicolumn{1}{c|}{Sub} & \multicolumn{1}{c|}{Del} & \multicolumn{1}{c|}{Ins} & WER \\ \hline
prune-single                  & \multirow{6}{*}{1-gram} & \multirow{7}{*}{1-gram} & \multicolumn{1}{c|}{5.2} & \multicolumn{1}{c|}{0.3} & \multicolumn{1}{c|}{1.2} & 6.8 \\  \cline{4-7} 
+ multi-level LM             &                         &                         & \multicolumn{1}{c|}{5.0}    & \multicolumn{1}{c|}{0.3}    & \multicolumn{1}{c|}{1.1}    &   6.4    \\ \cline{1-1} \cline{4-7} 

prune-recomb  & & & \multirow{2}{*}{4.8}  & \multirow{2}{*}{0.6}    & \multirow{2}{*}{0.7}    &   \multirow{2}{*}{6.1}   \\ 
+ multi-level LM & & & & & & \\ \cline{1-1} \cline{4-7} 

LF context approx               &                         &                         & \multicolumn{1}{c|}{5.0}    & \multicolumn{1}{c|}{0.7}    & \multicolumn{1}{c|}{0.8}    &   6.4   \\ \cline{1-1} \cline{4-7}
\multirow{3}{*}{N-best-list} &                         &                         & \multicolumn{1}{c|}{5.0}    & \multicolumn{1}{c|}{0.3}    & \multicolumn{1}{c|}{1.0}    &   6.3    \\ \cline{2-2} \cline{4-7}
                             & \multirow{2}{*}{4-gram} &  & \multicolumn{1}{c|}{4.7}    & \multicolumn{1}{c|}{0.5}    & \multicolumn{1}{c|}{0.6}    &    5.9   \\  \cline{3-7} 
                             &                   &   4-gram & \multicolumn{1}{c|}{4.6}    & \multicolumn{1}{c|}{0.5}    & \multicolumn{1}{c|}{0.6}    &   5.8    \\ \hline
\end{tabular}

\label{tab:wordhypgen}
\end{table}

\begin{table}[t!]
\centering
\caption{WER [\%] results of using the CE baseline transducer model and 1-/4-gram word-level LMs on LBS dev-other.}
\begin{tabular}{|c|cccc|}
\hline
\multirow{2}{*}{Recognition LM} & \multicolumn{4}{c|}{dev-other}                                                         \\ \cline{2-5} 
                    & \multicolumn{1}{c|}{Sub} & \multicolumn{1}{c|}{Del} & \multicolumn{1}{c|}{Ins} & WER \\ \hline
1-gram              & \multicolumn{1}{c|}{8.4} & \multicolumn{1}{c|}{1.5} & \multicolumn{1}{c|}{0.5} & 10.3  \\ \hline
4-gram              & \multicolumn{1}{c|}{4.7}    & \multicolumn{1}{c|}{0.6}    & \multicolumn{1}{c|}{0.6}    &   6.0    \\ \hline
\end{tabular}

\label{tab:unidecode}
\end{table}

\begin{table}[t!]
\centering
\caption{A subset of models from Table \ref{tab:wordhypgen}, but recognition with a word-level transformer LM on LBS dev-other and test-other.}
\setlength{\tabcolsep}{0.2em}
\begin{tabular}{|c|c|c|c|c|}
\hline

\multicolumn{2}{|c|}{Hyp-Space Generation} & Training & \multicolumn{2}{c|}{WER {[}\%{]}} \\ \cline{1-2}
Method & LM & LM & dev-other & test-other \\ \hline
prune-recomb & \multirow{2}{*}{1-gram}   & \multirow{3}{*}{1-gram} & \multicolumn{1}{c|}{3.9}       & 4.5   \\ \cline{1-1} \cline{4-5}
\multirow{3}{*}{N-best-list} &                           &                         & \multicolumn{1}{c|}{3.9}       & 4.6 \\ \cline{2-2} \cline{4-5} 
                             & \multirow{2}{*}{4-gram} &  & \multicolumn{1}{c|}{3.6}       & 4.2        \\ \cline{3-5} 
                             & & 4-gram & \multicolumn{1}{c|}{3.5}       & 4.1        \\ \hline
\end{tabular}
\label{tab:wordlfnbest}
\end{table}

\section{Conclusion}
In this paper, we investigate the effect of language models (LMs) with different context lengths and label units (phoneme vs. word) used in sequence discriminative training for phoneme-based neural transducers. For phoneme-level LMs, to apply a higher-order LM in a lattice-free (LF) manner, we introduce a context approximation method for the recombination step. This approximation is also applicable to word-level LMs, enabling the usage of word-level LMs in LF methods. Based on the proposed methods, we conduct a systematic comparison crossing N-best-list-based maximum mutual information (MMI) and minimum Bayes risk (MBR) training, as well as LF-MMI training. Experimental results on Librispeech consistently reveal the critical importance of the quality of the hypothesis space in sequence discriminative training. When the hypothesis space is of high quality, even using a small N-best list can achieve performance on par with employing all possible sequences as the hypothesis space. Comparing LMs with different contexts, we show that for both phoneme-level LMs and word-level LMs, the improvement obtained from a longer context is limited. Furthermore, we show that when the hypothesis space is well-constructed, using the word-level LM for MMI training yields better performance than using the phoneme-level LM, which indicates that the phoneme-based neural transducer can benefit from word-level statistics during training.
\section{Acknowledgement}
This work was partially supported by NeuroSys, which as part of the initiative “Clusters4Future” is funded by the Federal Ministry of Education and Research BMBF (03ZU1106DA), and by the project RESCALE within the program \textit{AI Lighthouse Projects for the Environment, Climate, Nature and Resources} funded by the Federal Ministry for the Environment, Nature Conservation, Nuclear Safety and Consumer Protection (BMUV), funding ID: 67KI32006A.

\bibliographystyle{IEEEbib}
\bibliography{strings,refs}

\end{document}